%% file: paper.tex
\documentclass{fairmeta}

\usepackage{amsmath}
\usepackage{amssymb}
\usepackage{wrapfig}
\usepackage{tikz}

\input{math_commands.tex}

\input{macros}

\crefname{section}{Section}{Sections}
\Crefname{section}{Section}{Sections}
\crefname{appendix}{Appendix}{Appendices}
\Crefname{appendix}{Appendix}{Appendices}
\crefname{figure}{Figure}{Figures}
\crefname{table}{Table}{Tables}
\crefname{equation}{Equation}{Equations}

\renewcommand\affiliationformat[2][]{{\small $^{#1}$#2}}

\title{\methodname{}: Connecting Task-Level Planning with Frame-Level Control for Game Agents}

\author[1\ast]{Guangzhi Wang}
\author[1\ast\dagger]{Junfu Pu}
\author[1\ast]{Yiran Wang}
\author[2\ast]{Xingyilang Yin}
\author[1]{Honghao Chen}
\author[1]{Cheng Cheng}
\author[3]{Lingen Li}
\author[4]{Yinan Zhou}
\author[3]{Jiarui Wu}
\author[5]{Yinuo Jiang}
\author[6]{Hongbo Jin}
\author[1,7\ddagger]{Xiaodong Cun}
\author[1]{Ying Shan}

\affiliation[1]{Tencent ARC Lab}
\affiliation[2]{University of Macau}
\affiliation[3]{The Chinese University of Hong Kong}
\affiliation[4]{City University of Hong Kong}
\affiliation[5]{Huazhong University of Science and Technology}
\affiliation[6]{Peking University}
\affiliation[7]{Great Bay University}

\contribution[\ast]{Equal contribution.}
\contribution[\dagger]{Project lead.}
\contribution[\ddagger]{Corresponding author.}

\abstract{
Building general game agents requires both high-level task reasoning and
low-latency action control in rapidly changing visual environments.
These capabilities favor models with substantially different computational
profiles, making them difficult to realize within a single model. In this work, we
present \methodname{}, a hierarchical planner--controller framework that connects
task-level planning with frame-level keyboard-and-mouse control. A
vision-language task planner decomposes a high-level goal into an ordered
sequence of atomic instructions and is optimized through supervised learning and
dependency-aware GRPO. The hidden states of special goal tokens that terminate each atomic
instruction provide continuous goal embeddings that are cached and reused across
control steps. A causal Transformer controller combines the embedding of the
active goal with recent observations to predict discrete keyboard and
mouse-button actions and continuous camera motion at every game step. We train
the system in three stages: planner training, controller pretraining, and joint alignment.
Experiments in Minecraft show that \methodname{} is competitive with
general-purpose vision-language models on instruction decomposition, and that
in closed-loop evaluation the agent achieves a higher combat success rate than the
evaluated baselines.
}

\date{\today}
\metadata[Project page]{\url{https://task-frame.github.io/}}

\begin{document}

\maketitle
\suppressfloats[t]

\input{sections/introduction}
\input{sections/related_work}
\input{sections/method}
\input{sections/experiments}
\input{sections/conclusion}

\clearpage
\bibliographystyle{assets/plainnat}
\bibliography{paper}

\clearpage
\beginappendix
\input{sections/appendix}

\end{document}

%% file: math_commands.tex
\usepackage{amsmath,amsfonts,bm}

\def\eqref#1{equation~\ref{#1}}

\def\1{\bm{1}}

\DeclareMathAlphabet{\mathsfit}{\encodingdefault}{\sfdefault}{m}{sl}
\SetMathAlphabet{\mathsfit}{bold}{\encodingdefault}{\sfdefault}{bx}{n}



%% file: macros.tex
\definecolor{pxk}{RGB}{28,28,28}
\definecolor{pxw}{RGB}{238,238,238}
\definecolor{pxb}{RGB}{96,64,42}
\definecolor{pxy}{RGB}{214,190,160}
\definecolor{pxp}{RGB}{240,164,176}
\definecolor{pxq}{RGB}{204,110,128}
\definecolor{pxg}{RGB}{86,150,84}
\definecolor{pxd}{RGB}{44,96,44}
\definecolor{pxs}{RGB}{128,128,128}
\definecolor{pxt}{RGB}{168,168,168}
\definecolor{pxu}{RGB}{92,92,92}
\newcommand{\pixicon}[1]{\begin{tikzpicture}[x=0.2em,y=0.2em,
    baseline={([yshift=-0.5ex]current bounding box.center)}]
  \foreach \row [count=\y] in {#1} {\foreach \c [count=\x] in \row
    {\fill[px\c] (\x,-\y) rectangle ++(1.04,1.04);}}
  \end{tikzpicture}}
\newcommand{\iconcow}{\pixicon{{b,b,b,b,b,b,b,b},{b,b,w,w,w,w,b,b},
  {w,k,w,w,w,w,k,w},{w,w,w,w,w,w,w,w},{b,y,y,y,y,y,y,b},{b,y,k,y,y,k,y,b},
  {b,y,y,y,y,y,y,b},{b,b,b,b,b,b,b,b}}}
\newcommand{\iconsheep}{\pixicon{{w,w,w,w,w,w,w,w},{w,w,w,w,w,w,w,w},
  {w,y,y,y,y,y,y,w},{w,y,k,y,y,k,y,w},{w,y,y,y,y,y,y,w},{w,y,y,p,p,y,y,w},
  {w,w,y,y,y,y,w,w},{w,w,w,w,w,w,w,w}}}
\newcommand{\iconpig}{\pixicon{{p,p,p,p,p,p,p,p},{p,p,p,p,p,p,p,p},
  {p,k,w,p,p,w,k,p},{p,p,p,p,p,p,p,p},{p,p,q,q,q,q,p,p},{p,q,k,q,q,k,q,p},
  {p,q,q,q,q,q,q,p},{p,p,p,p,p,p,p,p}}}
\newcommand{\iconzombie}{\pixicon{{d,d,d,d,d,d,d,d},{d,g,g,d,d,g,g,d},
  {g,g,g,g,g,g,g,g},{g,k,k,g,g,k,k,g},{g,g,g,d,d,g,g,g},{g,g,d,d,d,d,g,g},
  {g,d,g,g,g,g,d,g},{g,g,g,g,g,g,g,g}}}
\newcommand{\iconstone}{\pixicon{{s,s,t,s,s,s,u,s},{s,u,s,s,t,s,s,s},
  {t,s,s,u,s,s,s,t},{s,s,s,s,s,u,s,s},{s,t,s,s,s,s,t,s},{u,s,s,t,s,s,s,s},
  {s,s,u,s,s,t,s,u},{s,s,s,s,u,s,s,s}}}
\definecolor{pxo}{RGB}{216,175,147}
\definecolor{pxl}{RGB}{104,82,48}
\definecolor{pxm}{RGB}{156,122,74}
\newcommand{\iconiron}{\pixicon{{s,s,t,s,s,s,u,s},{s,o,o,s,t,s,s,s},
  {t,o,s,u,s,o,o,t},{s,s,s,s,s,o,s,s},{s,t,o,o,s,s,t,s},{u,s,o,t,s,s,o,s},
  {s,s,u,s,s,o,o,u},{s,s,s,s,u,s,s,s}}}
\newcommand{\iconcoal}{\pixicon{{s,s,t,s,s,s,u,s},{s,k,k,s,t,s,s,s},
  {t,k,s,u,s,k,k,t},{s,s,s,s,s,k,s,s},{s,t,k,k,s,s,t,s},{u,s,k,t,s,s,k,s},
  {s,s,u,s,s,k,k,u},{s,s,s,s,u,s,s,s}}}
\newcommand{\iconoak}{\pixicon{{l,m,l,l,m,l,l,m},{l,m,l,l,m,l,l,m},
  {m,l,l,m,l,l,m,l},{l,l,m,l,l,m,l,l},{l,m,l,l,m,l,l,m},{m,l,l,m,l,l,m,l},
  {l,l,m,l,l,m,l,l},{l,m,l,l,m,l,l,m}}}
\newcommand{\pmse}[1]{{\scriptsize\,$\pm$#1}}
\newcommand{\pmest}[1]{\est{\pmse{#1}}}
\newcommand{\taskhead}[2]{\shortstack{#1\\[1pt]{\scriptsize #2}}}

\newcommand{\methodname}{TaskFrame}

\newcommand{\est}[1]{#1}

%% file: sections/introduction.tex
\section{Introduction}
\label{sec:intro}

\begin{figure}[t]
    \centering
    \includegraphics[width=\linewidth]{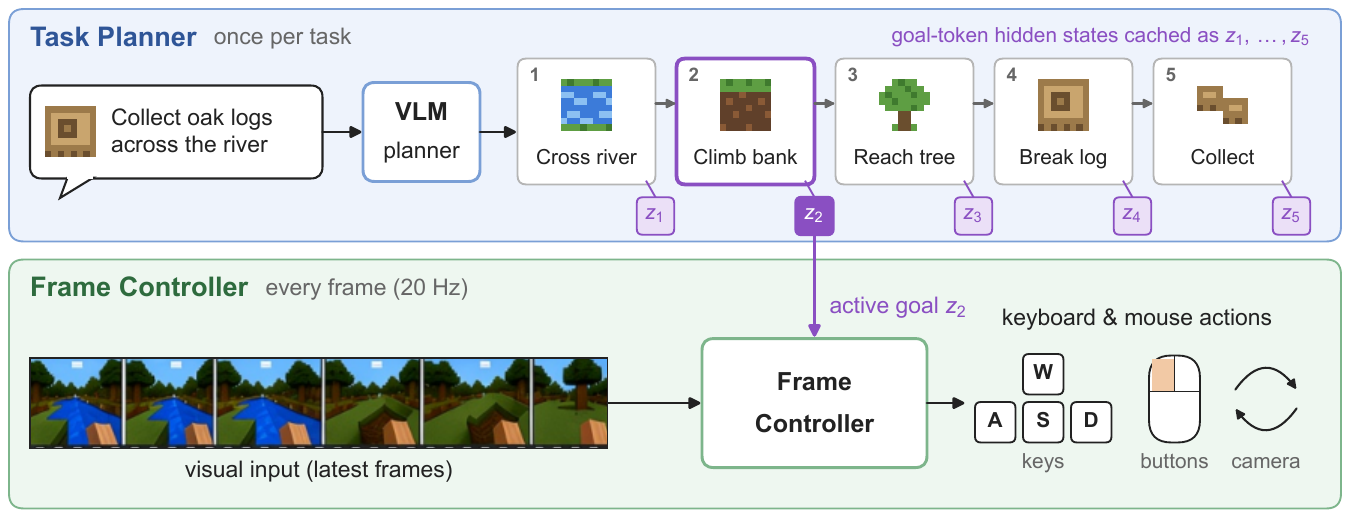}
    \caption{\textbf{Overview of \methodname{}.} Once per task, the task planner
    decomposes a high-level instruction into ordered atomic goals, each
    terminated by a special goal token whose hidden state is cached as a goal
    embedding. At every frame, the frame controller takes the embedding of the
    active goal and the latest visual input and emits native keyboard-and-mouse
    actions.}
    \label{fig:teaser}
\end{figure}

Games have long been a demanding testbed for artificial general intelligence
(AGI), as successful gameplay requires the integration of perception, reasoning,
planning, and decision making into goal-directed behavior.
Existing efforts have reached or surpassed expert human
performance in board games, Atari, multiplayer poker, and complex real-time
strategy games~\citep{silver2016alphago,silver2017alphagozero,
schrittwieser2020muzero,openai2019five,badia2020agent57,
brown2019pluribus,vinyals2019alphastar}. More recently, research has shifted
toward \emph{embodied game agents} that follow language instructions and act
directly from pixels in open-ended virtual worlds~\citep{sima2,nitrogen,openp2p,
jarvisvla,Hafner2025dreamer}. Unlike agents specialized for a fixed environment or
objective, such systems provide a natural setting for studying whether an agent
can understand diverse goals, adapt its behavior to changing visual situations,
and execute complex tasks through a general action interface.

Building such an embodied agent remains difficult because task-level reasoning
and frame-level control operate at different frequencies. Understanding a scene, grounding an
instruction, and determining the correct sequence of subgoals require a
large, high-capacity model. In contrast, executing those subgoals requires a
lightweight controller that can continuously incorporate the latest frame and
emit actions with low latency. This challenge is especially evident in visually
dynamic games, where abrupt camera motion and sudden events must be handled
through a hybrid action space combining discrete keyboard
presses with continuous mouse movement. Running a large reasoning model at every
control step is prohibitively expensive, whereas a small reactive policy alone
lacks the semantic capacity to reliably interpret and organize complex
instructions and visual history.

Existing work addresses parts of this tension through several strategies. Hierarchical agents use a language model to select skills or produce
high-level plans for a goal-conditioned controller~\citep{ahn2022saycan,
wang2023jarvis1,plan4mc,optimus2}. In physical robotics, dual-system Vision-Language-Action (VLA)
models separate a slower semantic module from a faster visuomotor policy~\citep{
gr00t,helix,hirt,robodual,hirobot}, while action chunking amortizes inference by predicting multiple
actions at once, with recent methods improving parallel decoding or asynchronous
execution~\citep{kim2025openvlaoft,black2025rtc}. A practical game agent must
connect these mechanisms through an interface that supports both structured
planning and responsive control. The high-level model must translate an
open-vocabulary goal into a complete and correctly ordered sequence of atomic
goals, and the controller must keep incorporating the latest visual observation
at every action step.

We present \textbf{\methodname{}}, a hierarchical game agent that connects
task-level planning with frame-level control through planner-produced goal
representations (\Cref{fig:teaser}). Its high-capacity \emph{task planner} is
trained to perform visually grounded instruction decomposition, with dedicated
data and a dependency-aware reinforcement learning objective that addresses both
step-count and ordering correctness. Each atomic goal it emits is terminated by a
special goal token, and the hidden states at these tokens are cached and passed
to a lightweight \emph{frame controller} as continuous goal embeddings. The
controller combines the embedding of the active goal with recent frames to emit
discrete keyboard and mouse-button actions and continuous camera motion at every
game step, so the large planner stays outside the frame-level control loop. We
train the system in three stages: the planner is trained for decomposition, the
controller is trained to produce actions from each frame, and the two modules are
finally aligned jointly.

We train the controller on human gameplay with native keyboard-and-mouse actions
recorded directly from players, and the planner on the hierarchical annotations
of GameHorizon Suite~\citep{wang2026gamehorizon}, which provide atomic
instructions, compositional goals, and dependency relations.
Our 9B planner outperforms GPT-4o and Qwen3.5-397B-A17B on instruction
decomposition. In closed-loop Minecraft evaluation, \methodname{} reaches a 51\%
average success rate on combat tasks, the highest among the evaluated models,
and ablations examine each component.
To summarize, our contributions are as follows:
\begin{itemize}
    \item We introduce \methodname{}, a hierarchical game agent in which a
    vision-language planner decomposes goals once per task and conditions a
    frame-level keyboard-and-mouse controller, which acts at every game step,
    through the hidden states of special goal tokens.
    \item We propose a dependency-aware GRPO objective for instruction
    decomposition that rewards necessary precedence relations separately from
    conventional ordering, derived from pairwise dependency annotations.
    \item We propose a three-stage training recipe that pretrains the controller
    with a generic goal encoder and then aligns it with the planner through LoRA
    adapters, and we equip the controller with a hybrid discrete and
    flow-matching action head. We evaluate the planner on two decomposition
    benchmarks and the full agent in closed-loop Minecraft control.
\end{itemize}

%% file: sections/related_work.tex
\section{Related Work}

\paragraph{Game agents.}
MineRL and VPT established large-scale imitation from human Minecraft
play~\citep{minerl,vpt}, and MineDojo added an open-ended environment with
internet-scale knowledge~\citep{minedojo}. STEVE-1, GROOT, ROCKET-1, and
MineDreamer condition low-level policies on text, reference videos,
visual-temporal prompts, or imagined goals~\citep{steve1,groot,rocket1,minedreamer},
while JARVIS-VLA and OmniJARVIS post-train VLMs to emit keyboard-and-mouse actions
directly~\citep{jarvisvla,omnijarvis}, following robotic
VLAs~\citep{RT2,openvla}. Recent systems generalize across tasks and
games~\citep{gametars,sima2,lumine,cradle} and control games through native
keyboard-and-mouse interfaces at frame rate~\citep{openp2p,nitrogen}. \methodname{}
instead connects explicit, trained task decomposition to frame-level control and
keeps the large VLM outside the per-frame loop.

\paragraph{Hierarchical task planning.}
Language-model planners ground decisions in skill affordances~\citep{ahn2022saycan},
use environment feedback~\citep{innermonologue}, or write code over control
primitives~\citep{codeaspolicies}. In Minecraft, Plan4MC, DEPS, Voyager, JARVIS-1,
and Optimus-1/2 plan over learned skills or dispatch multimodal plans to
goal-conditioned controllers~\citep{plan4mc,deps,voyager,wang2023jarvis1,
optimus1,optimus2}. These planners are prompted or supervised to produce plans
but are not optimized for whether their steps respect task dependencies.
Reinforcement learning with verifiable rewards, popularized by GRPO and its
variants~\citep{deepseekmath,deepseekr1,drgrpo}, offers a way to do so: we use the
must-precede annotations of GameHorizon Suite~\citep{wang2026gamehorizon} to define
a reward that separates necessary precedence from conventional ordering.

\paragraph{Dual-system control.}
Dual-system policies pair a slow semantic module with a fast visuomotor
policy~\citep{kahneman2011thinking}. Hi Robot and RT-H pass language between the
two~\citep{hirobot,rth}, and $\pi_{0.5}$ predicts a textual subtask before
decoding actions~\citep{pi05}; HiRT, RoboDual, GR00T N1, and Helix pass latent
features from a slowly updated VLM~\citep{hirt,robodual,gr00t,helix}, LCB trains a
dedicated latent token as the bridge~\citep{lcb}, and Acting While Understanding
updates semantic conditions asynchronously~\citep{actingwhileunderstanding}.
Action chunking, better action tokenization, and overlapping generation with
execution further improve control
efficiency~\citep{act,diffusionpolicy,pi0,fast,kim2025openvlaoft,black2025rtc,smolvla}.
Like HiRT and LCB, \methodname{} passes latent goals from a VLM to a fast policy,
but its planner is first optimized for ordered decomposition, and its latent
goals are the hidden states of goal tokens emitted within that decomposition, so
one planner pass yields both an interpretable plan and one embedding per step. 

%% file: sections/method.tex
\section{Method}
\label{sec:method}

\begin{figure*}[t]
  \centering
  \includegraphics[width=\linewidth]{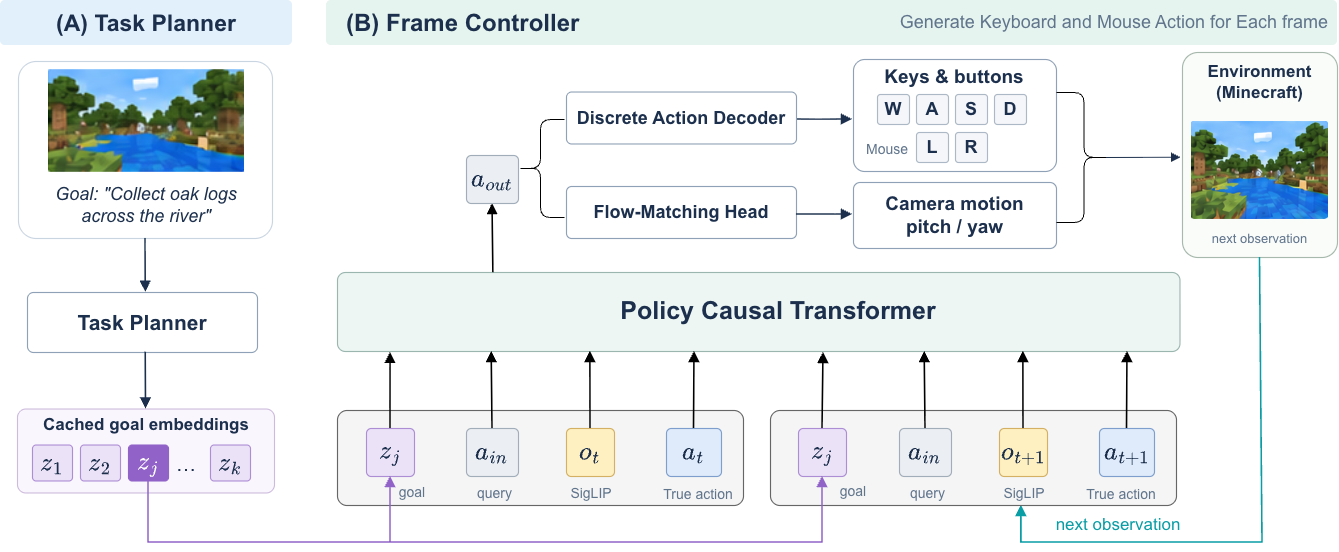}
  \caption{\textbf{Overview of \methodname{}}, a hierarchical planner--controller
  agent that separates task-level planning from frame-level control.
  \textbf{(A) Task Planner.} A vision--language model decomposes a high-level goal
  (e.g., \emph{``collect oak logs across the river''}) into an ordered sequence of
  atomic goals, each terminated by a special goal token. The hidden states at these
  tokens are cached as goal embeddings $z_1,\ldots,z_k$.
  \textbf{(B) Frame Controller.} A per-frame autoregressive policy. Each frame
  consumes the active goal embedding $z_j$, an action-query token $a_{\mathrm{in}}$, a SigLIP
  image token $o_t$, and the action $a_t$, which only later frames attend to. A causal
  Transformer aggregates them into the action-output token $a_{\mathrm{out}}$, which is
  decoded into discrete keyboard and mouse-button presses and continuous camera
  motion. The planner runs once per high-level goal, while the controller
  acts at every game step.}
  \label{fig:method}
\end{figure*}

\methodname{} connects task-level planning with frame-level game control
(\cref{fig:method}). A vision-language \emph{task planner} $\mathcal{B}$ maps an
observation and a high-level instruction to an ordered sequence of atomic goals,
each terminated by a special goal token $\tau$ whose hidden state serves as a
continuous representation of that goal. A causal \emph{frame controller}
$\mathcal{P}$ consumes the representation of the active goal together with its
recent frame and action history and predicts native keyboard-and-mouse actions at
every game step.

Training proceeds in three stages. First, we train the planner for visually
grounded instruction decomposition with supervised fine-tuning (SFT) and
dependency-aware GRPO (\cref{sec:brain}). Second, we pretrain the controller on
atomic-segment gameplay, using a frozen, generic vision-language model as its
goal encoder (\cref{sec:controller}). Third, we replace this encoder with the
trained planner and jointly adapt the controller and LoRA adapters in the planner
with the action objective (\cref{sec:alignment}). At control step $t$, the
controller observes a frame $o_t\in\mathbb{R}^{H\times W\times3}$ and predicts
\begin{equation}
  a_t=(a_t^{\mathrm{key}},a_t^{\mathrm{btn}},a_t^{\mathrm{mouse}}),
\end{equation}
where the first two components are discrete actions and the last is the
continuous camera motion produced by the mouse, represented as yaw and pitch
changes $a_t^{\mathrm{mouse}}=(\Delta\psi_t,\Delta\theta_t)\in\mathbb{R}^2$ in
degrees.

\subsection{Task Planner}
\label{sec:brain}




\paragraph{Visually grounded decomposition.}
The task planner $\mathcal{B}$ converts a visual observation and a high-level
instruction into an ordered sequence of atomic goals. Given an observation $o$
and a high-level instruction $g$, it generates
\begin{equation}
  \hat{\mathcal{S}}=(\hat{s}_1,\ldots,\hat{s}_{\hat{k}})
  \sim\pi_\theta(\cdot\mid o,g),
  \label{eq:brain-factorization}
\end{equation}
where each $\hat{s}_j$ is an atomic instruction. We train the planner on the hierarchical gameplay annotations of GameHorizon
Suite~\citep{wang2026gamehorizon}, which associate high-level goals with
temporally aligned atomic instructions and annotate precedence dependencies
between them. During training, the target number of goals $k$ is supplied through
a fixed reasoning prefix, yielding $\pi_\theta(\cdot\mid o,g,k)$; at inference it is
not supplied.

Starting from Qwen3.5-9B, we freeze the vision encoder and train the language model
and multimodal connector on the reference output sequence $\mathbf{y}$, whose goals
are each terminated by a goal token (\cref{eq:output-seq}):
\begin{equation}
  \mathcal{L}_{\mathrm{SFT}}
  =-\sum_{t=1}^{|\mathbf{y}|}
  \log\pi_\theta(y_t\mid o,g,k,y_{<t}).
  \label{eq:brain-sft}
\end{equation}
This stage teaches the planner to describe atomic goals in a consistent
decomposition format and provides the initial policy for subsequent refinement. Data statistics,
hyperparameters, and an analysis of why naive SFT overfits to annotation phrasing
are given in \cref{app:planner}.

\paragraph{Dependency-aware refinement.}
A reference plan records one execution sequence, although multiple orderings may
be valid. We refine the SFT policy with GRPO~\citep{deepseekmath,deepseekr1} using separate
rewards for step-count agreement and dependency-aware ordering.

For each input, generated steps are aligned one-to-one with reference steps
using Hungarian assignment. Pairwise similarity combines content-word F1 and a
character-level matching ratio:
\begin{equation}
  \operatorname{sim}(a,b)
  =0.5\,\mathrm{F1}_{\mathrm{content}}(a,b)
   +0.5\,\operatorname{ratio}_{\mathrm{char}}(a,b).
  \label{eq:step-similarity}
\end{equation}
Let $\operatorname{pos}[i]$ denote the generated position assigned to reference
step $i$. The count reward is
\begin{equation}
  R_k=\max\left(0,1-\frac{|\hat{k}-k|}{k}\right).
\end{equation}

We distinguish two ordering scores. \emph{Dependency satisfaction},
$S_{\mathrm{dep}}$, measures the fraction of annotated must-precede relations
respected by the generated plan. \emph{Conventional-order agreement},
$S_{\mathrm{conv}}$, measures agreement with the reference order among pairs
without an annotated dependency. The latter expresses a preference for the
recorded sequence rather than a requirement for task feasibility.

Let $D$ be the set of must-precede dependencies annotated in GameHorizon
Suite~\citep{wang2026gamehorizon} for a plan and
\begin{equation}
  F=\{(i,j):i<j,\ (i,j)\notin D,\ (j,i)\notin D\}.
\end{equation}
For a complete alignment,
\begin{align}
  S_{\mathrm{dep}}
  &=\frac{1}{|D|}\sum_{(i,j)\in D}
    \mathbf{1}[\operatorname{pos}[i]<\operatorname{pos}[j]],\\
  S_{\mathrm{conv}}
  &=\frac{1}{|F|}\sum_{(i,j)\in F}
    \mathbf{1}[\operatorname{pos}[i]<\operatorname{pos}[j]].
\end{align}
We combine these scores as
\begin{align}
  R_{\mathrm{ord}}
  &=\mathbf{1}[\hat{k}=k]
    \left(0.8S_{\mathrm{dep}}+0.2S_{\mathrm{conv}}\right),\\
  R&=0.3R_k+0.7R_{\mathrm{ord}}.
  \label{eq:brain-reward}
\end{align}
The soft count reward distinguishes different degrees of length mismatch. The
order reward requires the correct step count and a complete alignment, and
prioritizes necessary dependencies over conventional ordering. When one ordering
component is undefined, we use the available component alone. GRPO uses
mean-centered group rewards and a KL penalty to the SFT reference policy. The
0.8/0.2 split is derived from annotation statistics in \cref{app:planner-deps}.

\paragraph{Goal representations.}
Each atomic goal in the decomposition is terminated by a special goal token
$\tau$, so the output sequence is
\begin{equation}
  \mathbf{y}=
  [\operatorname{tok}(\hat{s}_1),\tau,\ldots,
   \operatorname{tok}(\hat{s}_{\hat{k}}),\tau].
  \label{eq:output-seq}
\end{equation}
Let $p_j$ be the position of the $j$-th goal token. The hidden state at this
position provides a continuous representation of goal $\hat{s}_j$,
\begin{equation}
  z_j=H_\theta(o,g,\mathbf{y})_{p_j},
  \label{eq:special-goal-state}
\end{equation}
which summarizes the goal in the context of the observation, the high-level
instruction, and the preceding steps. The decoded language supports
decomposition supervision and evaluation, while the controller consumes only
$z_j$.

\subsection{Frame Controller}
\label{sec:controller}
The frame controller is a per-frame autoregressive policy that emits native
keyboard-and-mouse actions from visual observations. We build on the
policy-transformer design of \citet{openp2p}, in which a causal Transformer reads a
per-frame token group and produces a single \emph{action-output} token that is
decoded into keyboard and mouse actions. We make two changes: the controller is
conditioned on the planner's vision-language goal embedding in addition to the
visual observation, and continuous camera motion is modeled with a flow-matching
head for smooth control.


\paragraph{Frame-level conditioning.}
Training samples are windows of $T$ consecutive frames, starting at $t_0$, that
lie within one annotated atomic segment. At frame $t$, the controller consumes the
goal embedding $z$ (\cref{sec:alignment}), a visual token $o_t$
obtained by feeding the frame into a frozen SigLIP encoder~\citep{siglip} followed
by a Perceiver resampler, a learned action-query token $a_{\mathrm{in}}$, and the
tokens of action $a_t$, which only later frames can attend to. The output of the
causal Transformer $f_\phi$ at the position of $a_{\mathrm{in}}$ is the
action-output token
\begin{equation}
  a^{\mathrm{out}}_t
  =f_\phi\!\left(z,\, o_{t_0:t},\, a_{t_0:t-1};\, a_{\mathrm{in}}\right),
  \qquad t_0\le t<t_0+T,
  \label{eq:controller-policy}
\end{equation}
from which the action heads decode $p_\phi(a_t\mid a^{\mathrm{out}}_t)$. The
previous actions $a_{t_0:t-1}$ are ground truth during training and the
controller's own outputs at inference.

\paragraph{Hybrid action head.}
The action-output token $a^{\mathrm{out}}_t$ is decoded by two heads. Discrete
keyboard keys and mouse buttons are produced by a small autoregressive action
decoder that samples the action tokens one at a time from
$a^{\mathrm{out}}_t$~\citep{openp2p}. Continuous camera motion is instead
generated by a lightweight flow-matching head~\citep{lipman2023flow,rectifiedflow},
which avoids discretizing small corrections and can represent several plausible
directions. The action objective is
\begin{equation}
  \mathcal{L}_{\mathrm{act}}
  =\mathcal{L}_{\mathrm{key}}
  +\mathcal{L}_{\mathrm{btn}}
  +\lambda_{\mathrm{mouse}}\mathcal{L}_{\mathrm{mouse}},
  \label{eq:action-loss}
\end{equation}
where $\mathcal{L}_{\mathrm{key}}$ and $\mathcal{L}_{\mathrm{btn}}$ are the
cross-entropy losses of the autoregressive decoder and
$\mathcal{L}_{\mathrm{mouse}}$ is the flow-matching loss.

\paragraph{Action conditioning and causal masking.}
Following \citet{openp2p}, the action-output token at frame $t$ cannot attend to
the same-frame action tokens, which would leak the target, and no token attends
to the action-output token; all other tokens are causal
(\cref{fig:attn-mask}).

\subsection{Planner--Controller Alignment}
\label{sec:alignment}
During controller pretraining (Stage 2), the goal embedding $z_0$ is produced by
a frozen, generic vision-language model $\mathcal{E}_0$ (Qwen3.5-9B) from the frame
at the start of the window and its atomic instruction $s$. In the third
stage, we replace $\mathcal{E}_0$ with the GRPO-trained planner $\mathcal{B}$,
augmented with LoRA adapters~\citep{hu2022lora} $\Delta$ in every Transformer block of its language
model. Each training window provides one annotated atomic goal $s$,
so the planner reads the first frame of the window and $s$, followed by the
goal token, and the hidden state at that token becomes the condition,
\begin{equation}
  z_{\mathcal{B}}
  =H_{\mathcal{B}+\Delta}\!\left([\,o_{t_0},\operatorname{tok}(s),\tau\,]\right)_{\tau},
  \label{eq:joint-condition}
\end{equation}
which replaces $z_0$ as the goal embedding $z$ in \cref{eq:controller-policy}. The
objective in
\cref{eq:action-loss} updates all controller parameters and $\Delta$ while the base
planner weights remain frozen. The resulting adapted planner is the single planner
used at inference, for both decomposition and goal embeddings. Additional
training details of this stage are given in \cref{app:controller-hparams}.


\subsection{Multi-Rate Inference}
\label{sec:inference}
Inference runs at two rates.
\emph{Task rate:} for a high-level goal $g$, the adapted planner decodes the plan
$(\hat{s}_1,\ldots,\hat{s}_{\hat{k}})$ once, without a supplied step count,
emitting a goal token after every atomic goal; the hidden states at these tokens
are cached as $(z_1,\ldots,z_{\hat{k}})$ (\cref{eq:special-goal-state}).
\emph{Frame rate:} the controller receives only the cached embedding of the
active goal, $z_j$, combines it with the frames and actions of the current
window, and emits an action at every game step. The decoded text is available
for inspection and evaluation but is not passed to the controller. 

%% file: sections/experiments.tex
\section{Experiments}
\label{sec:experiments}

We ask three questions: 1) Does \methodname{} act well in closed loop
(\cref{sec:closed-loop})? 2) Does its planner produce good plans
(\cref{sec:planner-exp})? 3) How should planning and control be connected
(\cref{sec:ablation})?

\subsection{Setup}
\label{sec:setup}

\paragraph{Training.}
The controller is trained on about 100 hours of VPT contractor
gameplay~\citep{vpt} with native keyboard-and-mouse actions, using the atomic
instructions of GameHorizon Suite~\citep{wang2026gamehorizon} as conditions.
Training windows contain $T=21$ frames (about one second at 20~Hz) and never
cross an atomic-segment boundary. The backbone is a 30-layer causal Transformer
of width 1024. The planner is trained on the decomposition annotations of GameHorizon Suite.

\paragraph{Closed-loop evaluation.}
We follow the programmatic protocol of JARVIS-VLA~\citep{jarvisvla} and
MCU~\citep{mcu} in MineStudio~\citep{minestudio} and evaluate two types of tasks:
\emph{mine} tasks place the equipped agent inside a chamber of the target block,
isolating block breaking from search, and \emph{kill} tasks spawn target animals a
few blocks ahead of an agent holding a sword. Episodes last at most 400 steps. We
compare with VPT-BC, VPT-RL~\citep{vpt}, STEVE-1~\citep{steve1}, and
JARVIS-VLA-7B~\citep{jarvisvla}.
Task configurations and baseline settings are given in
\cref{app:closed-loop}.

\paragraph{Planner evaluation.}
\emph{Soft Semantic F1} We utilize (SSF1) to measure step content on 2.1K held-out
plans. The planner sees one frame and the high-level instruction without the step
count, and the predicted and reference steps are matched by sentence-embedding
similarity, so paraphrases are accepted while missing or redundant steps are
penalized. \emph{Task2} of GameHorizon Suite~\citep{wang2026gamehorizon} measures
step ordering on 250 questions: given the step count, the planner generates a
plan, which is matched to the most similar of four candidate orderings; ties and
unparseable plans count as errors. More details are provided in
\cref{app:planner-eval}.

\subsection{Closed-Loop Control}
\label{sec:closed-loop}

\begin{table*}[t]
\centering
\small
\setlength{\tabcolsep}{4.8pt}
\caption{Closed-loop success rate (\%) on atomic Minecraft tasks over \est{3}
evaluation seeds, with 12 episodes per kill task and 6 per mining task in each
seed. Averages are reported with a binomial estimate of their standard deviation
across seeds; per-task mining results are in \cref{app:mine}.}
\label{tab:mc-main}
\setlength{\aboverulesep}{0pt}
\setlength{\belowrulesep}{0pt}
\renewcommand{\arraystretch}{1.35}
\begin{tabular}{l|ccccc|c}
\toprule
\multirow{2}{*}{Model} & \multicolumn{5}{c|}{\texttt{Kill Entities}}
& \texttt{Mine Blocks} \\
\cmidrule(lr){2-6}\cmidrule(lr){7-7}
& \taskhead{\iconcow}{Cow} & \taskhead{\iconsheep}{Sheep}
& \taskhead{\iconpig}{Pig} & \taskhead{\iconzombie}{Zombie} & {\scriptsize Avg.}
& \taskhead{\iconstone\,\iconiron\,\iconcoal\,\iconoak}{Avg.\ of 4} \\
\midrule
VPT-BC~\citep{vpt} & \est{17} & \est{17} & \est{17}
& \est{8} & \est{15}\pmest{5.4} & \est{38}\pmest{9.2} \\
VPT-RL~\citep{vpt} & \est{8} & \est{17} & \est{8}
& \est{0} & \est{8}\pmest{4.5} & \est{29}\pmest{8.8} \\
STEVE-1~\citep{steve1} & \est{17} & \est{8} & \est{17}
& \est{8} & \est{13}\pmest{5.2} & \est{62}\pmest{9.2} \\
JARVIS-VLA-7B~\citep{jarvisvla} & 17 & 33 & \textbf{50}
& 25 & 31\pmest{6.6} & \textbf{100}\pmest{5.3} \\
\midrule
\textbf{\methodname{} (Ours)} & \textbf{67} & \textbf{58}
& 42 & \textbf{35} & \textbf{51}\pmest{7.0} & 96\pmest{6.2} \\
\bottomrule
\end{tabular}
\end{table*}

\paragraph{Main results.}
\Cref{tab:mc-main} reports the results. \methodname{} reaches an average kill
success of 51\%, compared with 31\% for JARVIS-VLA-7B, and performs best on cow,
sheep, and zombie. \est{This gap is about twice the combined standard deviation.}
On mining, both \methodname{} and JARVIS-VLA-7B succeed in almost every episode,
so this category is close to saturated in our setting. \est{VPT-BC, VPT-RL, and STEVE-1 reach 8--15\% on kill
tasks and 29--62\% on mining.} Overall, \methodname{} achieves the best combat performance among the evaluated
models while remaining on par with JARVIS-VLA-7B on mining.

\begin{figure}[t]
\centering
\includegraphics[width=\linewidth]{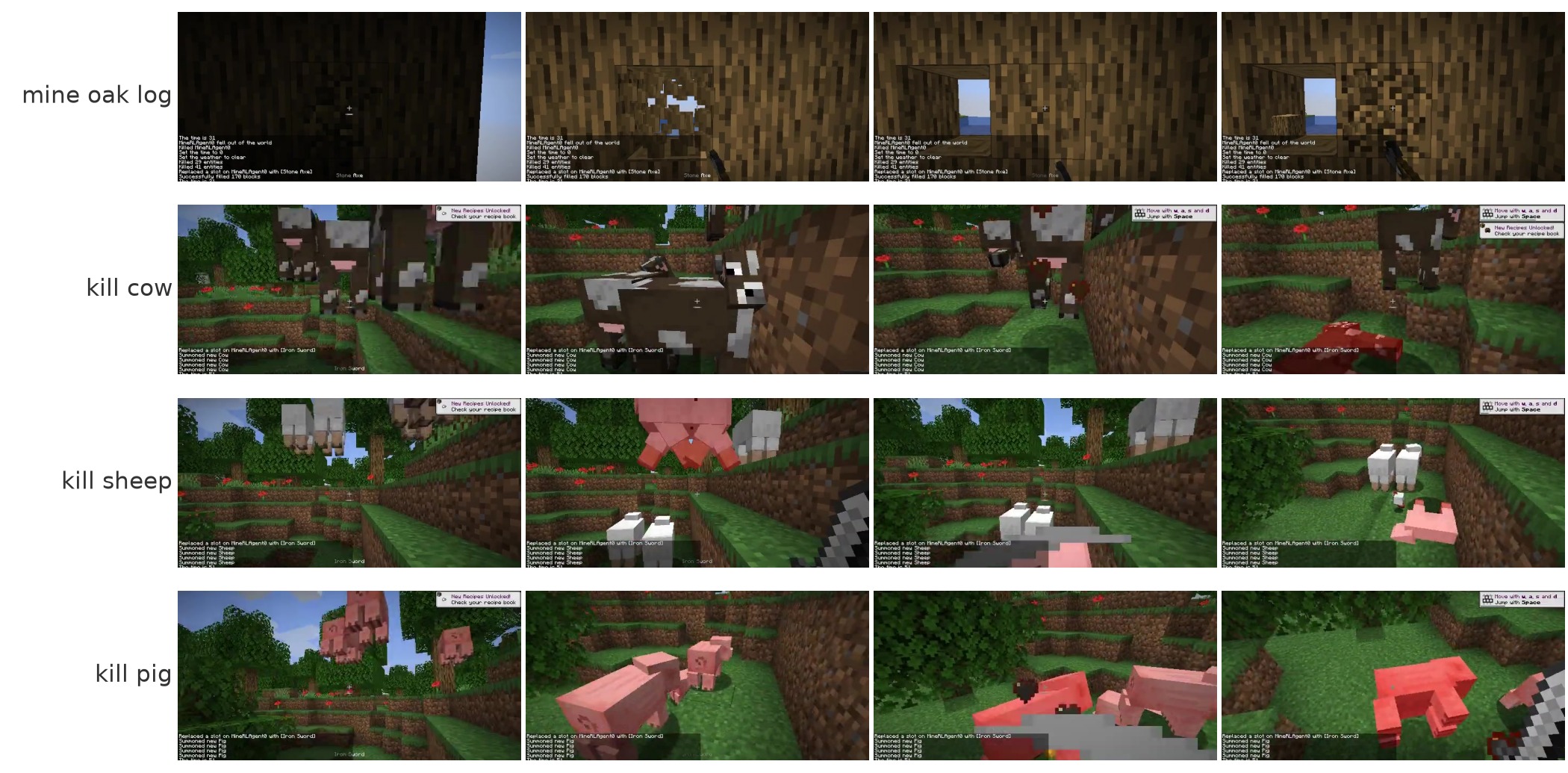}
\caption{\textbf{Closed-loop rollouts of \methodname{}.} Each row shows four frames of one successful episode on an atomic task. 
Comparisons with baselines are shown in \cref{fig:qual-compare}.}
\label{fig:qualitative}
\end{figure}

\paragraph{Qualitative results.}
\Cref{fig:qualitative} shows closed-loop rollouts. In combat, the controller turns
toward the target animals, approaches them through the grass, and keeps attacking
until the target is killed, even when several animals of different types are in
view; in mining, it repeatedly breaks the log in front of it.


\subsection{Task Planning}
\label{sec:planner-exp}

\begin{table}[t]
\centering
\small
\setlength{\tabcolsep}{8pt}
\caption{Planner performance. SSF1 is evaluated without a supplied step count,
whereas Task2 provides the target count. Unparsed is the percentage of questions without a valid decision.}
\label{tab:planner-main}
\setlength{\aboverulesep}{0pt}
\setlength{\belowrulesep}{0pt}
\renewcommand{\arraystretch}{1.22}
\begin{tabular}{l|c|cc}
\toprule
\multirow{2}{*}{Model} & \multirow{2}{*}{SSF1 $\uparrow$}
& \multicolumn{2}{c}{Task2} \\
\cmidrule(lr){3-4}
& & Acc. (\%) $\uparrow$ & Unparsed (\%) $\downarrow$ \\
\midrule
Gemini 3.5 Flash~\citep{gemini35flash} & \textbf{0.3815} & 44.4 & 13.6 \\
Gemini 3.6 Flash~\citep{gemini36flash} & 0.3808 & \textbf{46.8} & 15.2 \\
GPT-4o~\citep{gpt4o} & 0.2720 & 36.0 & 14.0 \\
Qwen3-Omni-30B-A3B~\citep{qwen3omni} & 0.2643 & 34.0 & 11.6 \\
Qwen3.5-397B-A17B~\citep{qwen35} & 0.2990 & 35.6 & 12.4 \\
Qwen3.6-35B-A3B~\citep{qwen36} & 0.2951 & 34.4 & \textbf{8.4} \\
BAGEL-7B-MoT~\citep{bagel} & -- & 28.0 & 14.4 \\
SenseNova-U1-8B-MoT~\citep{sensenova} & -- & 27.6 & 14.4 \\
\midrule
Qwen3.5-9B (backbone) & 0.2874 & 23.2 & 23.2 \\
\quad + SFT & 0.3312 & 38.8 & 14.0 \\
\quad + GRPO (\textbf{\methodname{} planner}) & 0.3459 & 40.0 & 11.6 \\
\bottomrule
\end{tabular}
\end{table}

As shown in \cref{tab:planner-main}, our 9B planner reaches an SSF1 of 0.3459 and
a Task2 accuracy of 40.0\%, exceeding GPT-4o and Qwen3.5-397B-A17B on both metrics
while remaining below the two Gemini models, which are evaluated zero-shot like
the other general-purpose baselines. Over the backbone, SFT raises SSF1 from
0.2874 to 0.3312 and Task2 accuracy from 23.2\% to 38.8\%, and GRPO further
improves both metrics and lowers the Unparsed rate from 14.0\% to 11.6\%. The gain
in SSF1, which uses sentence embeddings rather than the reward's step matcher and
supplies no step count, indicates that GRPO does not merely fit its own reward.

\subsection{Connecting Planning and Control}
\label{sec:ablation}

\paragraph{Planner and controller roles.}
To isolate the value of the hierarchy and of joint alignment, we predict actions
on held-out gameplay with high-level goals (\cref{tab:abl-roles}; setup in
\cref{app:roles}). \emph{Planner only} generates actions directly with the
planner; \emph{controller only} uses an empty goal; \emph{cascade} passes the
planner's decoded instructions as text to the Stage-2 controller; \methodname{}
passes the aligned planner's goal embeddings; and \emph{oracle goals} uses
annotated atomic instructions as an upper bound.

\begin{table}[t]
\centering
\small
\setlength{\aboverulesep}{0pt}
\setlength{\belowrulesep}{0pt}
\renewcommand{\arraystretch}{1.25}
\begin{minipage}[t]{0.57\linewidth}
\centering
\caption{Offline action prediction on held-out gameplay with high-level goals.}
\label{tab:abl-roles}
\setlength{\tabcolsep}{3pt}
\begin{tabular}{l|ccc}
\toprule
Method & Key F1 $\uparrow$ & Btn.\ F1 $\uparrow$ & Cam.\ RMSE $\downarrow$ \\
\midrule
Planner only & \est{0.21} & \est{0.38} & \est{6.6} \\
Controller only & \est{0.16} & \est{0.32} & \est{6.8} \\
Cascade (text) & \est{0.24} & \est{0.45} & \est{5.9} \\
\textbf{\methodname{}} & \est{\textbf{0.27}} & \est{\textbf{0.49}}
& \est{\textbf{5.6}} \\
\midrule
\textcolor{gray}{Oracle goals} & \est{0.29} & \est{0.52} & \est{5.4} \\
\bottomrule
\end{tabular}
\end{minipage}\hfill
\begin{minipage}[t]{0.40\linewidth}
\centering
\caption{Closed-loop kill success (\%) when only the supplied goal is changed.}
\label{tab:abl-goal-main}
\setlength{\tabcolsep}{4pt}
\begin{tabular}{l|c}
\toprule
Goal & Kill avg.\ $\uparrow$ \\
\midrule
\textbf{\methodname{}} (full) & \textbf{51}\pmest{3.5} \\
Empty goal & \est{25}\pmest{3.0} \\
Mismatched goal & \est{13}\pmest{2.3} \\
Planner w/o observation & \est{31}\pmest{3.1} \\
\bottomrule
\end{tabular}
\end{minipage}
\end{table}

\est{The planner alone beats the unconditioned controller, but it emits actions
for a whole window from a single observation and cannot react within the window;
the controller alone cannot tell which behavior is intended. The cascade recovers
most of the gap, and the
aligned goal embeddings of \methodname{} give the best results on all metrics};
since both use the same decomposition, \est{the gap between them isolates the
effect of joint alignment. Oracle goals add little, so most remaining error lies in the
controller rather than in the decomposition.}

\paragraph{Goal conditioning in closed loop.}
We keep the trained model fixed and change only the goal it receives, evaluating
each setting on the kill tasks with \est{48} episodes per task in each of three
seeds (\cref{tab:abl-goal-main}; per-task results in \cref{app:goal}). \est{With an
empty goal, kill success drops from 51\% to 25\%; some success remains because
every spawned animal is a valid target. A mismatched goal lowers it further to
13\%, so the controller actively follows the goal rather than ignoring it, and
withholding the observation from the planner costs 20 points}, showing that the
visual grounding of the goal helps.

\begin{wraptable}{r}{0.38\linewidth}
\vspace{-1.2em}
\centering
\small
\setlength{\tabcolsep}{5pt}
\setlength{\aboverulesep}{0pt}
\setlength{\belowrulesep}{0pt}
\renewcommand{\arraystretch}{1.25}
\caption{Camera MAE ($^\circ$) on held-out gameplay.}
\label{tab:abl-mouse}
\begin{tabular}{l|cc}
\toprule
Camera head & Yaw $\downarrow$ & Pitch $\downarrow$ \\
\midrule
Discrete bins & \est{2.1} & \est{1.2} \\
Flow matching & \est{\textbf{1.7}} & \est{\textbf{1.0}} \\
\bottomrule
\end{tabular}
\vspace{-1em}
\end{wraptable}

\paragraph{Camera head.}
We compare the flow-matching camera head with the discrete head of
\citet{openp2p}, which classifies camera motion into bins and takes their
expected value. The two controllers differ only in the camera head and are
trained with the same data and schedule (\cref{tab:abl-mouse}; setup in
\cref{app:mouse}). \est{Flow matching lowers the mean absolute error from 2.1$^\circ$ to
1.7$^\circ$ for yaw and from 1.2$^\circ$ to 1.0$^\circ$ for pitch.} Binning rounds small corrections to the nearest bin and averages over
several plausible directions, which hurts most when aiming at a moving target.

%% file: sections/conclusion.tex
\section{Conclusion}
\label{sec:conclusion}

We presented \methodname{}, a hierarchical game agent that separates task-level
planning from frame-level keyboard-and-mouse control. A vision-language planner, trained with supervised
decomposition and dependency-aware GRPO, turns a high-level instruction into an
ordered sequence of atomic goals and outperforms
general-purpose models such as GPT-4o and Qwen3.5-397B-A17B on instruction
decomposition. The hidden states of the goal tokens in its plan serve as goal
embeddings. A causal Transformer controller with a hybrid action head converts the active embedding and
recent frames into native actions at every game step. In closed-loop Minecraft
evaluation, the agent achieves the highest combat success rate among all evaluated
models. \est{Ablations further show that the controller relies on the goal condition and
that the aligned goal embeddings are more effective than passing decoded text
between the two modules.}

\paragraph{Limitations.}
The agent does not decide on its own when an atomic goal is done, so it relies
on an external execution protocol to switch to the next goal, and it does not
replan after a failure. In addition, the controller sees only about one second
of history, which is short for long-horizon tasks.
Finally, all experiments are conducted in Minecraft, and extending the method to
other games remains a challenging direction for future work.

%% file: sections/appendix.tex
\section{Task Planner Details}
\label{app:planner}

\subsection{Decomposition data}
\label{app:planner-data}
We use the Minecraft decomposition corpus of GameHorizon
Suite~\citep{wang2026gamehorizon}; its construction, annotation procedure, and
statistics are described there. We split it by complete video clip, yielding
57,901 training examples from 1,119 clips and 2,200 held-out examples from 50
disjoint clips; the held-out set is the SSF1 evaluation set of
\cref{sec:experiments}. After adding the second description style and the
paraphrased instructions and downsampling thematic examples without replacement,
the SFT training set contains 40,068 examples: 10,442 causal, 2,914
original-segment, and 26,712 thematic.

\subsection{Evaluation protocols}
\label{app:planner-eval}
\paragraph{Soft Semantic F1.}
The model receives one observation frame and a high-level instruction without
the reference step count. Predicted and reference steps are encoded with
\texttt{all-mpnet-base-v2}, and pairs with cosine similarity below $0.65$ are
excluded before alignment. Causal examples use order-preserving one-to-one
alignment, whereas thematic examples use unrestricted Hungarian assignment. For
$n$ reference steps, $m$ predicted steps, and matched pairs $\mathcal{M}$,
\begin{equation}
  \mathrm{SSF1}
  =\frac{2\sum_{(i,j)\in\mathcal{M}}\cos(e_i,\hat{e}_j)}{n+m}.
  \label{eq:ssf1}
\end{equation}
We report the mean over the 2,200 held-out examples; unparseable outputs receive
zero.

\paragraph{Task2.}
We use the high-level-to-atomic decomposition task (Task2) of GameHorizon
Suite~\citep{wang2026gamehorizon} with a generate-and-align protocol on 250
Minecraft questions. The model receives one observation frame, a high-level
instruction, and the target step count, and generates a numbered plan. Each
question provides four permutations of the same canonical step set. Generated
steps are aligned with the canonical steps using the similarity in
\cref{eq:step-similarity}, the aligned sequence is compared with each candidate by
Kendall correlation, and the unique highest-scoring candidate is selected. The
Unparsed rate is the percentage of questions without a valid decision, including
ties; these count as incorrect.

\subsection{Training hyperparameters}
\label{app:planner-hparams}

\begin{table}[h]
\centering
\small
\caption{Planner training hyperparameters.}
\label{tab:planner-hparams}
\setlength{\aboverulesep}{0pt}
\setlength{\belowrulesep}{0pt}
\renewcommand{\arraystretch}{1.2}
\begin{tabular}{l|l|l}
\toprule
 & SFT & GRPO \\
\midrule
Initialization & Qwen3.5-9B & SFT checkpoint \\
Trainable modules & language model, connector & language model, connector \\
Training data & 40,068 examples, 1 epoch & 2,207 examples \\
Peak learning rate & $2\times10^{-5}$ & $1\times10^{-6}$ \\
Schedule & cosine, 0.5\% warmup & cosine, 3\% warmup \\
Weight decay & 0.01 & -- \\
Batch & 4 per GPU $\times$ 8 GPUs & 32 generations per step \\
Generations per input & -- & 8 \\
KL coefficient $\beta$ & -- & 0.04 \\
Max.\ sequence / generation length & 8,192 & 512 \\
Gradient clipping & -- & 1.0 \\
Parallelism & ZeRO-3 & ZeRO-3, colocated inference (TP=2) \\
\bottomrule
\end{tabular}
\end{table}

The vision encoder is frozen in both stages. Following \citet{drgrpo}, GRPO uses
mean-centered group rewards without dividing by the within-group standard
deviation: when all
candidates in a group score similarly, a very small standard deviation would
amplify negligible reward differences into large advantages, whereas unnormalized
rewards preserve the intended relative scale of the count and order terms.

\subsection{Dependency annotations and reward weights}
\label{app:planner-deps}
The dependency annotations of GameHorizon Suite~\citep{wang2026gamehorizon} show
that causal plans are much more constrained than thematic ones, and that a large
fraction of thematic plans are essentially unordered, so a single canonical-order
reward would over-constrain the thematic majority.

To set the weights of \cref{eq:brain-reward}, we simulate the value of each
ordering signal on the annotated graphs. A uniformly random permutation reaches
19.8\% on the Task2 ordering criterion. Satisfying all annotated dependencies adds
30.8 points, whereas additionally following the recorded order on the remaining
free pairs adds only 4.6 points. The implied contributions of dependency and
conventional ordering are approximately 87:13, which motivates the 0.8/0.2 split
between $S_{\mathrm{dep}}$ and $S_{\mathrm{conv}}$.

\subsection{GRPO training dynamics}
\label{app:planner-grpo}

\begin{table}[h]
\centering
\small
\caption{GRPO training statistics over the first 450 steps (means within each
window).}
\label{tab:grpo-dynamics}
\setlength{\aboverulesep}{0pt}
\setlength{\belowrulesep}{0pt}
\renewcommand{\arraystretch}{1.2}
\begin{tabular}{l|ccccc}
\toprule
Steps & Total reward & Count score & Order score & KL & Grad.\ norm \\
\midrule
1--50 & 0.313 & 0.630 & 0.178 & 0.003 & 1.669 \\
51--150 & 0.331 & 0.664 & 0.188 & 0.021 & 1.619 \\
151--250 & 0.356 & 0.694 & 0.211 & 0.050 & 1.408 \\
251--350 & \textbf{0.379} & 0.702 & \textbf{0.240} & 0.044 & 1.496 \\
351--450 & 0.369 & \textbf{0.704} & 0.226 & 0.040 & 1.150 \\
\bottomrule
\end{tabular}
\end{table}

The count score rises from 0.630 to 0.704, and the order score improves from
0.178 to 0.226--0.240 before plateauing. KL to the SFT policy stays between 0.02
and 0.05 after warmup, and the gradient norm decreases without signs of policy
collapse. Generated plans average 73--86 tokens and never reach the 512-token
limit, so count errors are not truncation artifacts.

\subsection{Why naive SFT generalizes poorly}
\label{app:planner-sft}
Our first SFT run balanced categories by duplicating minority examples. Its
training loss decreased steadily, but on an external diagnostic built from
VPT-20h (539 four-way questions over 498 clips that share no recordings with our
training data; chance is 25\%), accuracy fell to 39.5\%, below the backbone's
51.2\%. Grouping questions by how closely the gold steps match the phrasing of
the training annotations shows that the SFT model beats the backbone on the most
familiar quartile (76.1\% vs.\ 70.1\%) but falls below chance on the least
familiar one (17.0\% vs.\ 35.6\%). The model thus learns a narrow decomposition
skill tied to annotation phrasing, which a clip-level holdout from the training
distribution did not expose. These findings motivated our SFT recipe
(\cref{app:planner-data}): we balance categories by downsampling thematic
examples instead of duplicating minority ones, describe each atomic target in two
styles, and paraphrase high-level instructions.

\section{Frame Controller Details}
\label{app:controller}

\subsection{Training data}
\label{app:controller-data}
The controller is trained on the VPT-100h gameplay of GameHorizon
Suite~\citep{wang2026gamehorizon}, about 100 hours of first-person Minecraft play
by human contractors~\citep{vpt} with per-frame keyboard, mouse, and camera
actions and atomic-instruction annotations; we use its 1,132 training clips and
35 validation clips at 20~fps and refer to it for data collection and annotation
details. Frames are resized from $640\times360$ to $256\times256$ without cropping
to preserve the full field of view, and GUI frames (inventory and crafting
screens) are excluded. We split each clip into \emph{clean segments} in which the instruction
is constant and no GUI frame occurs, giving 33,392 segments; training windows are
tiled inside a single segment, which removes the roughly 16\% of windows that
would otherwise straddle an instruction boundary.

\begin{figure}[h]
\centering
\includegraphics[width=0.49\linewidth]{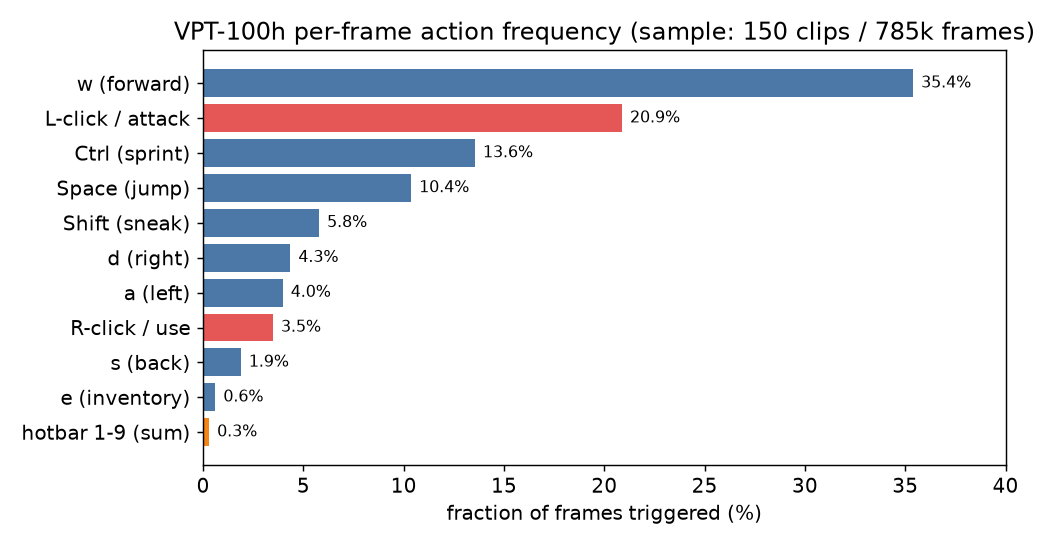}\hfill
\includegraphics[width=0.49\linewidth]{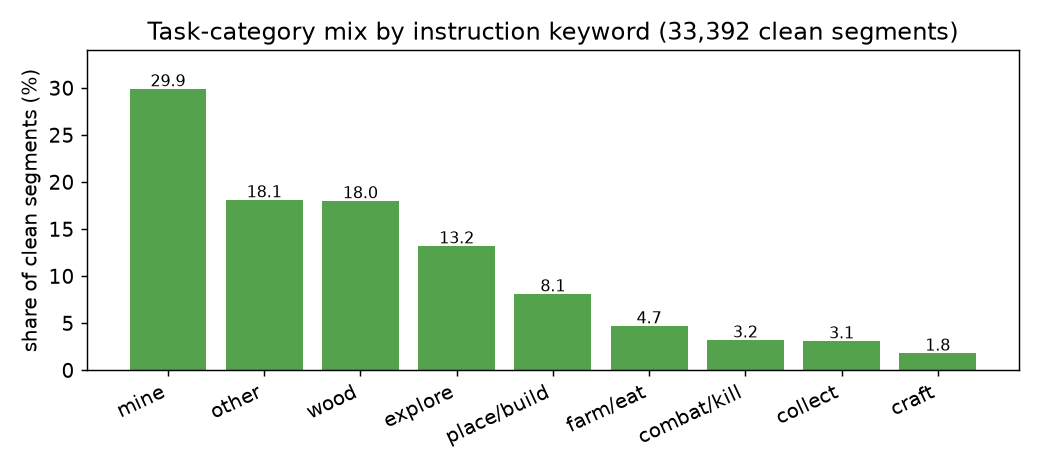}
\caption{Statistics of the controller training data. \textbf{Left:} per-frame
action frequency on a sample of 150 clips (785K frames). \textbf{Right:} task
category mix of clean segments, grouped by instruction keyword.}
\label{fig:vpt-stats}
\end{figure}

The data is strongly long-tailed (\cref{fig:vpt-stats}). Moving forward occurs in
35.4\% of frames and attacking in 20.9\%, whereas using or placing items occurs in
3.5\% and hotbar selection in only 0.3\%; 54.7\% of frames contain no key press.
Combat accounts for only 3.2\% of clean segments. We therefore oversample segments
by category and by rare actions: combat segments by a factor of 6, placement and
building by 3, and segments containing a use action by 1.5.

\subsection{Action space}
\label{app:action-space}
The keyboard head predicts up to four simultaneous keys from 26 classes: no key,
W, A, S, D, space (jump), left and right shift (sneak), left control (sprint), E
(inventory), Q (drop), F (swap hands), Z, the four arrow keys, and hotbar slots
1--9. The mouse-button head predicts up to two simultaneous buttons from four
classes (none, attack, use, middle). Both are decoded autoregressively from the
action-output token. The flow-matching head predicts camera motion as yaw and
pitch increments in degrees, converted from recorded pixel displacement at
0.15$^\circ$ per pixel and normalized by per-axis standard deviations estimated
from the data (about 5$^\circ$ for yaw and 2$^\circ$ for pitch). Camera targets are
masked on GUI frames.

\subsection{Architecture and optimization}
\label{app:controller-hparams}

\begin{table}[h]
\centering
\small
\caption{Frame controller hyperparameters.}
\label{tab:controller-hparams}
\setlength{\aboverulesep}{0pt}
\setlength{\belowrulesep}{0pt}
\renewcommand{\arraystretch}{1.2}
\begin{tabular}{l|l}
\toprule
Backbone & causal Transformer, 30 layers, width 1024, 16 heads \\
Initialization & pretrained policy of \citet{openp2p}, layers duplicated \\
 & to 30 with near-identity initialization \\
Context & $T=21$ frames ($\approx$1\,s), attention window 21 frames \\
Visual encoder & frozen SigLIP, $256\times256$ input; 256 patches compressed \\
 & to 1 token per frame by a Perceiver resampler \\
Goal condition & hidden state at the goal token, MLP projection \\
Stage-2 goal encoder & frozen Qwen3.5-9B, current frame and the instruction \\
Stage-3 planner & GRPO planner (Qwen3.5-9B) with LoRA rank 64, $\alpha=128$, \\
 & learning rate $5\times10^{-6}$ \\
Keyboard / button decoder & autoregressive, 26 / 4 classes \\
Mouse head & flow matching, hidden size 512, 10 Euler steps \\
Loss weights & $\lambda_{\mathrm{mouse}}=1.0$ \\
Optimizer & AdamW, $\beta=(0.9,0.999)$, weight decay $10^{-4}$ \\
Learning rate & $5\times10^{-5}$, 500 warmup steps \\
Batch & 2 per GPU $\times$ 16 GPUs, bf16 mixed precision \\
Gradient clipping & 1.0 \\
\bottomrule
\end{tabular}
\end{table}

\paragraph{Auxiliary losses.}
\est{During joint alignment (Stage 3), the adapted planner is additionally trained
to describe the actions taken in the current window as a short sequence of action
phrases, with a cross-entropy loss (weight 1.5) that up-weights rare action
categories such as camera, interaction, and hotbar actions.}
The reported controller was additionally trained with an auxiliary
video-prediction loss (weight 0.1): a branch initialized from
Wan2.1-T2V-1.3B~\citep{wan} predicted 8 future frames from the controller's
hidden states and was discarded at inference, so it does not affect the policy's
architecture or inference cost. In our experiments, removing this loss had little
effect on performance, and we do not consider it part of the method.

\begin{figure}[h]
\centering
\includegraphics[width=0.6\linewidth]{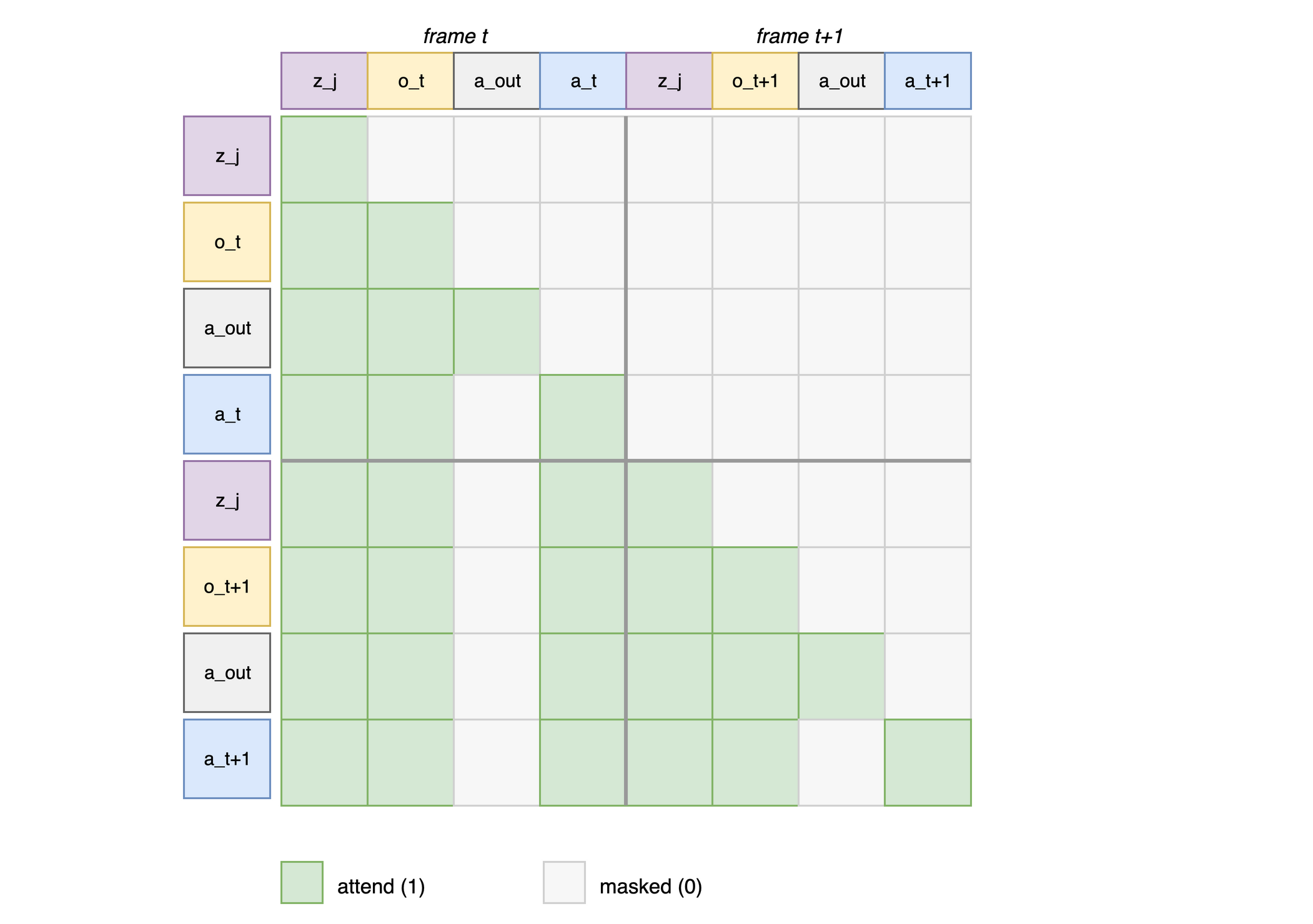}
\caption{Causal attention mask of the frame controller, adopted
from~\citet{openp2p} and shown for two consecutive frames. Rows are query tokens
and columns are key tokens; green cells are attended and gray cells are masked.
Each frame contributes the goal condition $z_j$, the visual token $o_t$, the
action-output token $a^{\mathrm{out}}$, and the ground-truth action $a_t$. No token
attends to $a^{\mathrm{out}}$, and $a^{\mathrm{out}}$ at frame $t$ attends to
previous ground-truth actions but not to the same-frame action $a_t$.}
\label{fig:attn-mask}
\end{figure}

\subsection{Flow-matching camera head}
\label{app:mouse}
We compare the flow-matching camera head with a discrete head that classifies
camera motion into bins and predicts their expected value, following
\citet{openp2p}. The two controllers are identical except for the camera head,
and both are trained with the same data and schedule and evaluated on the held-out
VPT-100h validation split; results are in \cref{tab:abl-mouse}.

\section{Closed-Loop Evaluation Details}
\label{app:closed-loop}

\subsection{Task configurations}
\label{app:tasks}
All tasks run in MineStudio~\citep{minestudio} with $256\times256$ observations
for \methodname{}, a 400-step limit, and success defined as the first positive
reward from the corresponding event. \emph{Mine} tasks equip the agent with the
required pickaxe or axe and fill a hollow $7\times5\times7$ shell of the target
block around it, so a target is visible and reachable from any direction; the
instruction is, for example, ``Mine stone blocks using a pickaxe.'' \emph{Kill}
tasks equip an iron sword and spawn four target animals 2--6 blocks in front of
the agent, with the instruction ``Attack and kill the cow with a sword.''; the
zombie task is run at night with night vision. Task configurations are frozen
and shared across all compared models.

\subsection{Baseline settings}
\label{app:baselines}
\emph{JARVIS-VLA-7B} is served with vLLM and decoded with its own tokenized action
space and $\mu$-law camera discretization at $640\times360$, using the sampling
temperature of its official rollout (0.9). \emph{UI-TARS-1.5-7B}, which appears
only in the qualitative comparison of \cref{app:qual}, receives $640\times360$
observations with its native interleaved screenshot-and-action history, and each
decision is held for 20 game steps because the model emits GUI events at well
below 1~Hz. Its prompt describes the effect of each action without suggesting a
strategy, and we sample with temperature 0.7, because greedy decoding on
near-identical consecutive frames repeats the same action indefinitely. Its GUI
events (click, drag, hotkey) are mapped to game actions by a fixed adapter.

\subsection{Per-task mining results}
\label{app:mine}
\Cref{tab:mc-mine-full} reports the per-task mining results summarized by the
Mine column of \cref{tab:mc-main}. \methodname{} and JARVIS-VLA-7B succeed in
almost every episode, so these tasks do not separate the two models.

\begin{table}[h]
\centering
\small
\setlength{\tabcolsep}{5pt}
\caption{Per-task closed-loop success rate (\%) on mining tasks over \est{3}
evaluation seeds (6 episodes per seed), with binomial estimates of the standard
deviation across seeds.}
\label{tab:mc-mine-full}
\setlength{\aboverulesep}{0pt}
\setlength{\belowrulesep}{0pt}
\renewcommand{\arraystretch}{1.35}
\begin{tabular}{l|ccccc}
\toprule
\multirow{2}{*}{Model} & \multicolumn{5}{c}{\texttt{Mine Blocks}} \\
\cmidrule(lr){2-6}
& \taskhead{\iconstone}{Stone} & \taskhead{\iconiron}{Iron ore}
& \taskhead{\iconcoal}{Coal ore} & \taskhead{\iconoak}{Oak log} & {\scriptsize Avg.} \\
\midrule
VPT-BC~\citep{vpt} & \est{67}\pmest{19.6} & \est{17}\pmest{16.8} & \est{17}\pmest{16.8}
& \est{50}\pmest{20.4} & \est{38}\pmest{9.2} \\
VPT-RL~\citep{vpt} & \est{50}\pmest{20.4} & \est{0}\pmest{10.5} & \est{17}\pmest{16.8}
& \est{50}\pmest{20.4} & \est{29}\pmest{8.8} \\
STEVE-1~\citep{steve1} & \est{83}\pmest{16.8} & \est{33}\pmest{19.6} & \est{50}\pmest{20.4}
& \est{83}\pmest{16.8} & \est{62}\pmest{9.2} \\
JARVIS-VLA-7B~\citep{jarvisvla} & \textbf{100}\pmest{10.5} & \textbf{100}\pmest{10.5}
& \textbf{100}\pmest{10.5} & \textbf{100}\pmest{10.5} & \textbf{100}\pmest{5.3} \\
\midrule
\textbf{\methodname{} (Ours)} & \textbf{100}\pmest{10.5} & \textbf{100}\pmest{10.5}
& 83\pmest{16.8} & \textbf{100}\pmest{10.5} & 96\pmest{6.2} \\
\bottomrule
\end{tabular}
\end{table}

\subsection{Qualitative comparison with baselines}
\label{app:qual}
\begin{figure}[h]
\centering
\includegraphics[width=0.92\linewidth]{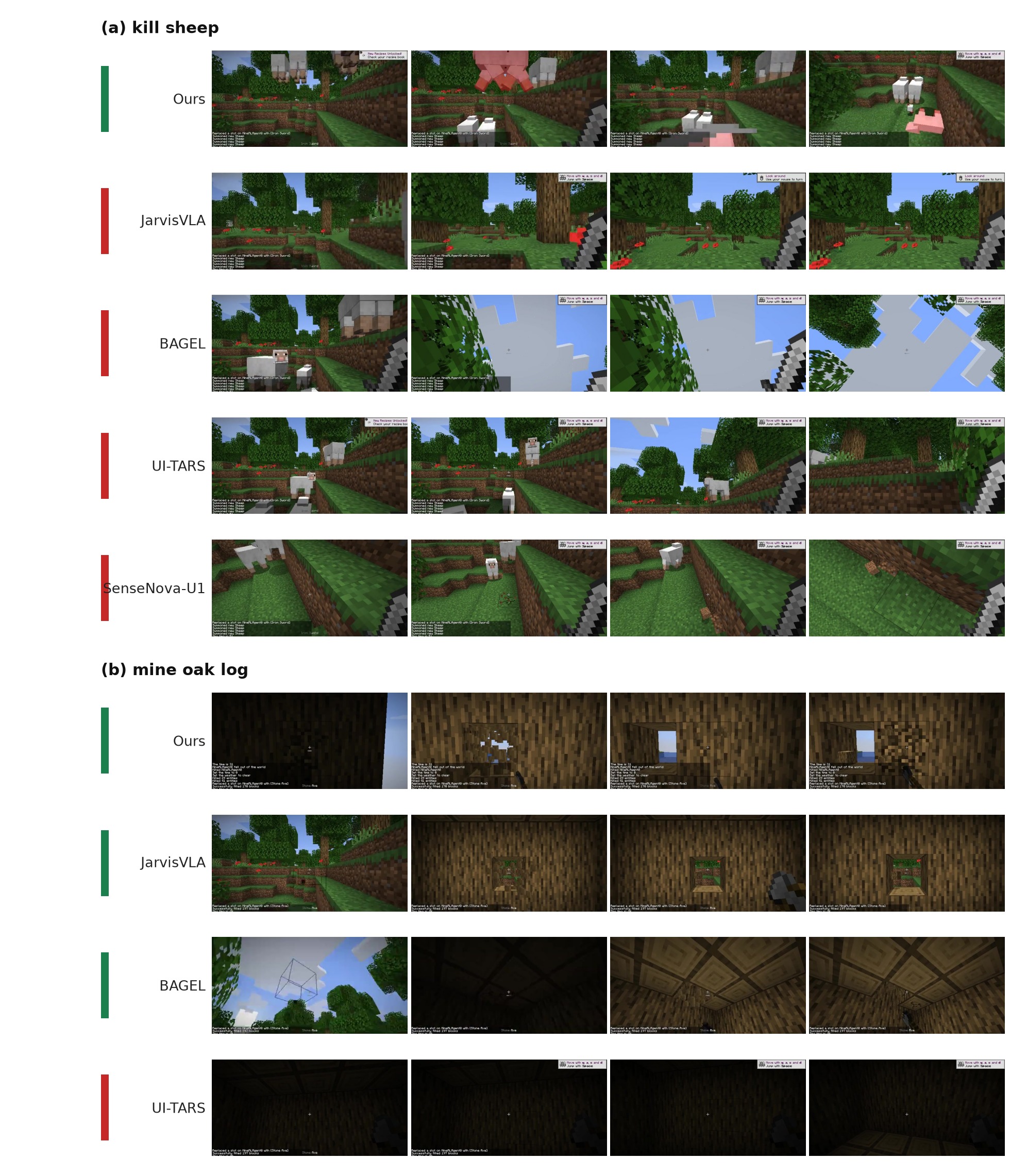}
\caption{\textbf{Qualitative comparison on (a) kill sheep and (b) mine oak log.}
Each row shows four frames of one episode, from left to right. Green and red bars
indicate success and failure.}
\label{fig:qual-compare}
\end{figure}

\Cref{fig:qual-compare} compares rollouts of \methodname{} with those of the
baselines on the same scenes; green and red bars mark successful and failed
episodes. We additionally include the unified multimodal models
BAGEL~\citep{bagel} and SenseNova-U1~\citep{sensenova}, run zero-shot through the
same action adapter as UI-TARS-1.5-7B. On kill sheep, only \methodname{}
succeeds: JARVIS-VLA-7B wanders without engaging the sheep, BAGEL turns the camera
toward the sky and does not recover, UI-TARS-1.5-7B turns away from the targets,
and SenseNova-U1 approaches a sheep but does not kill it. Mining oak logs is
easier: \methodname{}, JARVIS-VLA-7B, and BAGEL succeed, whereas UI-TARS-1.5-7B
remains in front of the wall without breaking it.

\subsection{Goal-conditioning ablation}
\label{app:goal}
We keep the trained model fixed and change only the goal it receives: an empty
instruction (\emph{controller only}), the instruction of a different task such as
mining in a combat scene (\emph{mismatched goal}), or a goal embedding computed
without showing the image to the planner (\emph{planner w/o observation}).
\Cref{tab:abl-goal} gives the per-task results summarized in
\cref{tab:abl-goal-main}.

\begin{table}[h]
\centering
\small
\setlength{\tabcolsep}{4pt}
\caption{Closed-loop goal-conditioning ablations on kill tasks (success rate,
\%, over \est{3} seeds with \est{48} episodes per task each; $\pm$ values are
binomial estimates of the standard deviation across seeds). The model is unchanged; only the goal supplied at inference
differs.}
\label{tab:abl-goal}
\setlength{\aboverulesep}{0pt}
\setlength{\belowrulesep}{0pt}
\renewcommand{\arraystretch}{1.3}
\begin{tabular}{l|ccccc}
\toprule
Setting & \taskhead{\iconcow}{Cow} & \taskhead{\iconsheep}{Sheep}
& \taskhead{\iconpig}{Pig} & \taskhead{\iconzombie}{Zombie} & {\scriptsize Avg.} \\
\midrule
\methodname{} (full) & \textbf{67}\pmest{6.8} & \textbf{58}\pmest{7.1}
& \textbf{42}\pmest{7.1} & \textbf{35}\pmest{6.9} & \textbf{51}\pmest{3.5} \\
\midrule
Controller only (empty goal) & \est{42}\pmest{7.1} & \est{33}\pmest{6.8}
& \est{17}\pmest{5.4} & \est{8}\pmest{3.9} & \est{25}\pmest{3.0} \\
Mismatched goal & \est{25}\pmest{6.3} & \est{17}\pmest{5.4} & \est{8}\pmest{3.9}
& \est{0}\pmest{1.5} & \est{13}\pmest{2.3} \\
Planner w/o observation & \est{50}\pmest{7.2} & \est{42}\pmest{7.1}
& \est{25}\pmest{6.3} & \est{8}\pmest{3.9} & \est{31}\pmest{3.1} \\
\bottomrule
\end{tabular}
\end{table}

\subsection{Offline action-prediction setup}
\label{app:roles}
The comparison in \cref{tab:abl-roles} uses \est{50} clips that are disjoint from
both planner and controller training, each annotated with a high-level goal and
its temporally aligned atomic segments. The high-level goal and the first frame
are given to the system, atomic goals are advanced at the annotated segment
boundaries, and actions are predicted at every frame given the ground-truth
history and compared with the recorded actions. \emph{Planner only} fine-tunes
the planner with LoRA to generate a compact action sequence for each window from
frames and the active instruction. \emph{Cascade} passes the planner's decoded
atomic instructions as text to the Stage-2 controller, whose frozen generic
encoder never saw planner outputs. We report macro-F1 over key classes and mouse
buttons and camera RMSE in degrees.

\subsection{Original JARVIS-VLA configurations}
\label{app:orig}
We also evaluate \texttt{kill\_zombie} and \texttt{mine\_stone} with the original
JARVIS-VLA task configurations (\cref{tab:mc-orig}). The zombie task spawns a
single zombie at night and equips the agent with a diamond sword. The stone task
spawns the agent at a random location without placing stone nearby, so the agent
must first find exposed stone.

\begin{table}[h]
\centering
\small
\setlength{\tabcolsep}{6pt}
\caption{Success rate (\%) under the original JARVIS-VLA task configurations
over \est{3} seeds, with binomial estimates of the standard deviation across
seeds (12 episodes per task in each seed).}
\label{tab:mc-orig}
\setlength{\aboverulesep}{0pt}
\setlength{\belowrulesep}{0pt}
\renewcommand{\arraystretch}{1.35}
\begin{tabular}{l|cc}
\toprule
Model & \taskhead{\iconzombie}{Kill zombie} & \taskhead{\iconstone}{Mine stone} \\
\midrule
JARVIS-VLA-7B & 42\pmest{14.3} & \textbf{17}\pmest{11.4} \\
\midrule
\methodname{} (Ours) & \textbf{83}\pmest{11.4} & 0\pmest{5.6} \\
\bottomrule
\end{tabular}
\end{table}

\methodname{} doubles the zombie success of JARVIS-VLA-7B but fails the original
stone task, whereas JARVIS-VLA-7B
succeeds in 17\% of episodes. Unlike the mining scenes of \cref{tab:mc-main},
this task requires locating stone from a random spawn before mining, a search
behavior that extends well beyond the controller's one-second context and is
rare in atomic training segments. We therefore attribute the gap to exploration
rather than to block breaking, which \methodname{} performs reliably once the
target is in view.